\documentclass[conference]{IEEEtran}
\IEEEoverridecommandlockouts

\usepackage{cite}
\usepackage{amsmath,amssymb,amsfonts}
\usepackage{algorithmic}
\usepackage{graphicx}
\usepackage{textcomp}
\usepackage{xcolor}
\def\BibTeX{{\rm B\kern-.05em{\sc i\kern-.025em b}\kern-.08em
    T\kern-.1667em\lower.7ex\hbox{E}\kern-.125emX}}
\begin{document}


\title{Facial Expression-based Parkinson's Disease Severity Diagnosis via Feature Fusion and Adaptive Class Balancing
\vspace{-4mm}}


\author{
\IEEEauthorblockN{Yintao Zhou}
\IEEEauthorblockA{\textit{School of Mathematics and Computer Sciences} \\
\textit{Nanchang University}\\
Nanchang, China \\
yintaozhou@email.ncu.edu.cn}
\and
\IEEEauthorblockN{Wei Huang}
\IEEEauthorblockA{\textit{Yichun University} \\
Yichun, China \\
n060101@e.ntu.edu.sg}
\and
\IEEEauthorblockN{Zhengyu Li}
\IEEEauthorblockA{\textit{Nanchang University Second Affiliated Hospital} \\
\textit{Nanchang University}\\
Nanchang, China \\
antcarrot@126.com}
\and
\IEEEauthorblockN{Jing Huang}
\IEEEauthorblockA{\textit{Nanchang University Second Affiliated Hospital} \\
\textit{Nanchang University}\\
Nanchang, China \\
413007230067@email.ncu.edu.cn}
\and
\IEEEauthorblockN{Meng Pang$^*$}
\IEEEauthorblockA{\textit{School of Mathematics and Computer Sciences} \\
\textit{Nanchang University}\\
Nanchang, China \\
mengpang@ncu.edu.cn}
}
\maketitle
\begin{abstract}
Parkinson's disease (PD) severity diagnosis is crucial for early detecting potential patients and adopting tailored interventions.
Diagnosing PD based on facial expression is grounded in PD patients' ``masked face'' symptom and gains growing interest recently for its convenience and affordability.
However, current facial expression-based approaches often rely on single type of expression which can lead to misdiagnosis, and ignore the class imbalance across different PD stages which degrades the prediction performance. 
Moreover, most existing methods focus on binary classification (i.e., PD / non-PD) rather than diagnosing the severity of PD.
To address these issues, we propose a new facial expression-based method for PD severity diagnosis which integrates multiple facial expression features through attention-based feature fusion. 
Moreover, we mitigate the class imbalance problem via an adaptive class balancing strategy which dynamically adjusts the contribution of training samples based on their class distribution and classification difficulty.
Experimental results demonstrate the promising performance of the proposed method for PD severity diagnosis, as well as the efficacy of attention-based feature fusion and adaptive class balancing.
\end{abstract}


\vspace{-4mm}
\section{Introduction}
\vspace{-1mm}

Parkinson's disease (PD) is a prevalent neurodegenerative disorder causing motor symptoms (e.g., rigidity, bradykinesia, and postural instability) and non-motor symptoms (e.g., insomnia, hyposmia, and cognitive impairment), severely impairing patients' life quality~\cite{bloem2021parkinson}.
With the aging global population, PD imposes significant burdens on healthcare systems and social economy,
affecting over 8.05 million people worldwide---a number projected to double within 30 years~\cite{huang2023global}.
Although PD is incurable currently, severity diagnosis is critical for timely and targeted intervention to slow disease progression~\cite{bloem2021parkinson}.

\begin{figure}[!t]
\setlength{\abovecaptionskip}{-3pt}
\centerline{\includegraphics[width=2.8in]{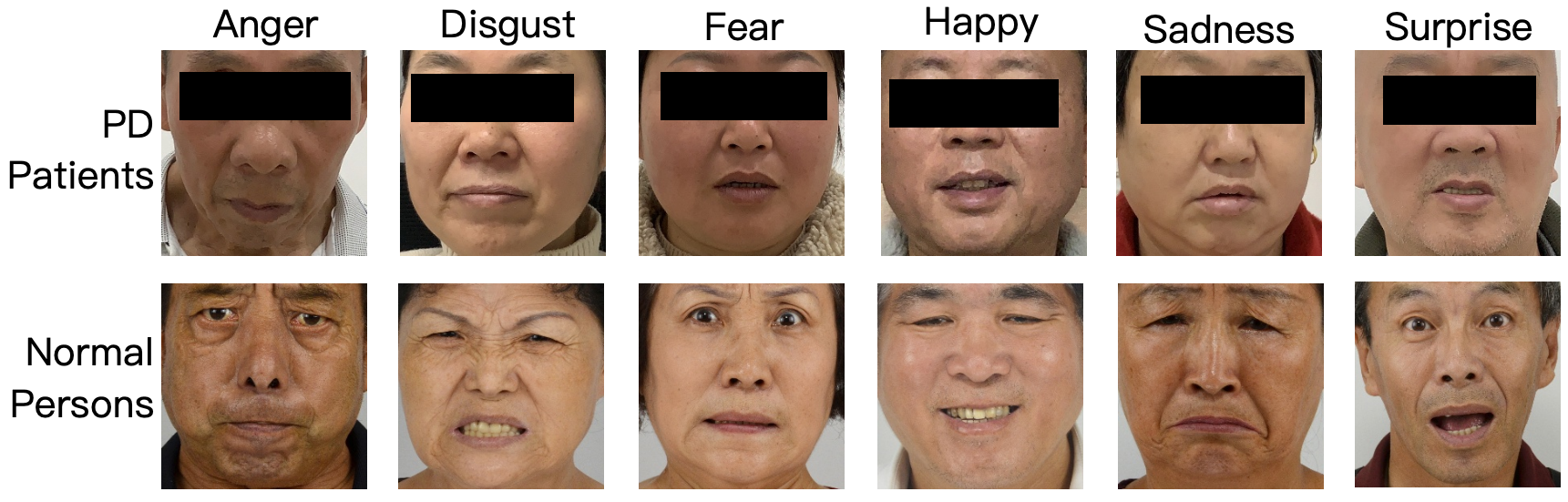}}
\caption{A comparison of facial expressions between PD patients and normal persons, with the patients' eye regions cut out for privacy protection.}
\label{fig1}
\vspace{-10mm}
\end{figure}

Existing methods for PD diagnosis fall into two types~\cite{huang2022facial}:
1) in-vivo methods based on medical imaging techniques (e.g., MRI, PET, and SPECT) with high accuracy but limited by their reliance on specialized equipments and prohibitive costs;
2) in-vitro methods based on the collection and analysis of in-vitro biomarkers including gait signal~\cite{meng2025inertial} and speech signal~\cite{he2024exploiting}.
Recently facial expression~\cite{huang2022facial,bandini2017analysis,pang2025unified} has emerged as a promising PD biomarker for its cross-culture universality and convenient data acquisition.
The motivation of facial expression-based PD diagnosis comes from the clinical observation that most PD patients exhibit the ``masked face'' symptom characterized by reduced facial muscle movement~\cite{huang2022facial}, depicted in Fig.~\ref{fig1}.
However, current approaches face two constraints:
1) relying on single type of expression such as smiling and ignoring that some PD patients can still make certain types of expressions at the early stage~\cite{bandini2017analysis};
2) the imbalanced class distribution across different PD stages can interfere with model training and degrade the diagnosis performance.

To address the above issues, we propose a new facial expression-based method for PD severity diagnosis by: 1) integrating features from 6 basic emotional facial images (i.e., happiness, sadness, surprise, fear, anger, and disgust) through attention-based feature fusion to reduce misdiagnosis;
2) designing an adaptive class balancing strategy to dynamically adjust samples' contribution based on their class distribution and classification difficulty during training, so as to mitigate bias caused by class imbalance across different PD stages.
Experiments on the PD facial expression dataset demonstrate the superiority and effectiveness of the proposed method.

\vspace{-3mm}
\section{Proposed Method}
\vspace{-0.5mm}

\begin{figure}[!t]
\setlength{\abovecaptionskip}{-1pt}
\centerline{\includegraphics[width=.48\textwidth]{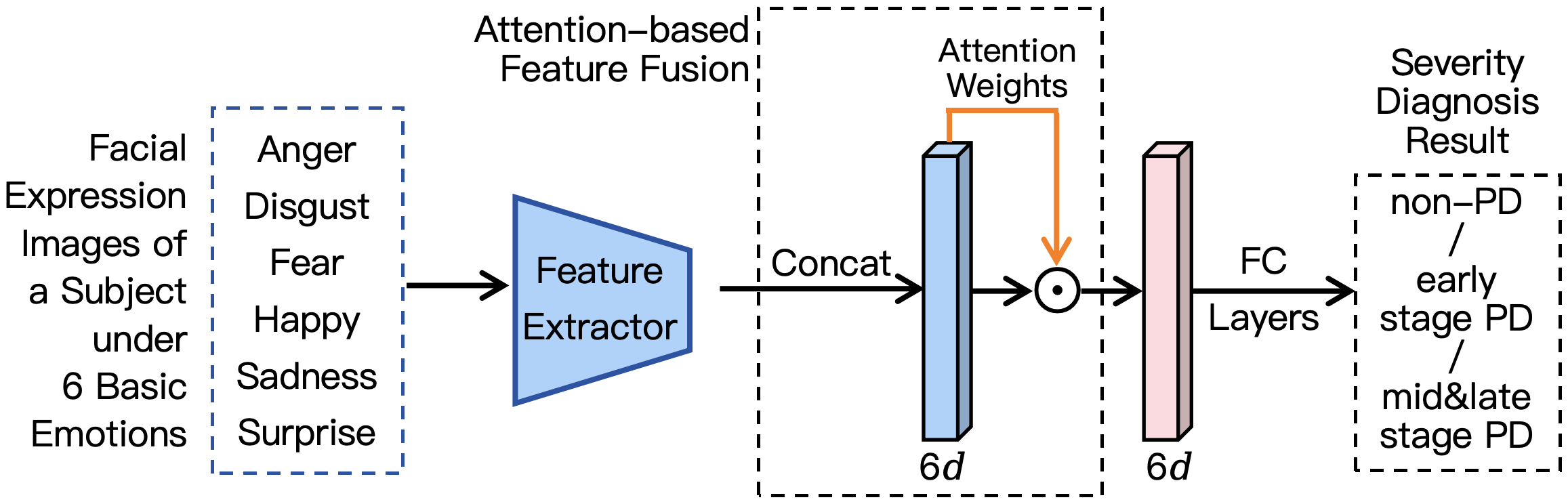}}
\caption{The pipeline of the proposed severity diagnosis method for PD.}
\label{fig2}
\vspace{-6mm}
\end{figure}

The proposed PD severity diagnosis method illustrated in Fig.~\ref{fig2} takes facial expression images of a subject under 6 basic emotions as input.
Then, facial expression features are obtained from each image by the feature extractor which employs a ResNet-18~\cite{he2016deep} pretrained for expression recognition.

\textbf{Attention-based Feature Fusion}: 
The features extracted from six expression images are then integrated via attention-based feature fusion:
we concatenate the six $d$-dimensional facial expression features along the channel dimension, followed by pooling operations and fully connected (FC) layers to calculate the attention weights which strengthen important feature channels while suppressing irrelevant ones, thereby achieving effective feature fusion. 
This process can be formulated as~\eqref{eq1}:
\vspace{-2mm}
\begin{equation}
W_{attn} = \sigma ( FC ( AvgPool(F)) + FC ( MaxPool(F))),
\label{eq1}
\vspace{-1mm}
\end{equation}

\noindent{where $F$ denotes the concatenated $6d$-dimensional feature, $FC(\cdot)$ denotes FC layers, and $AvgPool(\cdot)$ / $MaxPool(\cdot)$ are average / max pooling operations.}
$\sigma$ is the sigmoid function which maps input into the interval of $[0,1]$.
The attention weights $W_{attn}$ are then element-wise multiplied by $F$ to obtain the fused feature, which is then transformed by two FC layers into the result vector indicating model's prediction for PD severity, i.e., non-PD / early-stage PD / mid\&late-stage PD.

\textbf{Adaptive Class Balancing}:
During training, we design an adaptive class balancing strategy which adaptively adjusts the contribution of samples based on their class distribution and classification difficulty, so as to mitigate the performance degradation brought by imbalanced sample distribution across different PD stages.
The corresponding adaptive focal loss for a given sample is formulated as~\eqref{eq3}:
\vspace{-1.5mm}
\begin{equation}
L_{adapt} = -\sum_{i=1}^{C} \alpha_i (1 - p_t)^{\gamma} y_i \log(p_i),
\label{eq3}
\vspace{-1.5mm}
\end{equation}
where $C$ is the total number of classes.
$\alpha_i$ represents the adaptive class weighting factor determined by the class frequency ratio: $\alpha_i=N_{max}/N_i$, where $N_{i}$ and $N_{max}$ denote the sample count of the $i$-th class and the most frequent class, respectively.
This ensures that samples of minority class receive higher weights, while those of majority class receive lower weights, thereby balancing classes of different size.
The term $(1 - p_t)^{\gamma}$ represents the modulating factor aiming to emphasize hard, misclassified samples while down-weighting well-classified ones, 
with $p_t$ denoting model's predicted probability of the sample belonging to its true class and $\gamma$ denoting the focusing parameter.
$y_i$ and $p_i$ represent the $i$-th PD severity label and the predicted  probability of this class, respectively.
\vspace{-1mm}
\section{Experimental Results}


\textbf{Evaluation on PD Severity Diagnosis:}
See the Appendix for dataset and implementation details.
Table~\ref{tab1} presents the performance of our method and three popular deep learning approaches for image classification including MobileNetV4~\cite{qin2024mobilenetv4}, Swin-Transformer~\cite{liu2021swin}, and EfficientNetV2~\cite{tan2021efficientnetv2}, along with two ablated versions of our method: one without attention-based feature fusion (``Ours w/o AFF") and another without adaptive class balancing (``Ours w/o ACB").
The results demonstrate the superiority of our method for PD severity diagnosis, as well as the effectiveness of attention-based feature fusion and adaptive class balancing.

\begin{table}[!t]
\caption{Performance Comparison for PD Severity Diagnosis.}
\vspace{-4mm}
\begin{center}
\begin{tabular}{|c|c|c|c|c|}
\hline
\textbf{Methods} & \textbf{AUC} & \textbf{G-Mean} & \textbf{F1-Score} & \textbf{Acc} \\
\hline
MobileNetV4~\cite{qin2024mobilenetv4} & 0.8545 & 0.3814 & 0.6644 & 0.7849 \\

Swin-Transformer~\cite{liu2021swin} & 0.7954 & 0.0238 & 0.2594 & 0.6369 \\

EfficientNetV2~\cite{tan2021efficientnetv2} & 0.8574 & 0.5086 & 0.6221 & 0.7565 \\
\hline
Ours w/o AFF & 0.8696 & 0.5402 & 0.6507 & 0.8197 \\

Ours w/o ACB & 0.8717 & 0.3318 & 0.6093 & 0.8293 \\
\hline
Ours & \textbf{0.8816} & \textbf{0.5750} & \textbf{0.6718} & \textbf{0.8354}\\
\hline
\end{tabular}
\label{tab1}
\end{center}
\vspace{-7mm}
\end{table}

\vspace{-1mm}
\section{Conclusion}
\vspace{-1mm}

In this paper we propose a new facial expression-based PD severity diagnosis method using attention-based feature fusion to reduce misdiagnosis and adaptive class balancing to handle imbalanced data distribution. Experiments demonstrate its superiority in PD severity diagnosis.
\vspace{-1mm}
\section*{Acknowledgment}
\vspace{-1mm}
This work is supported in part by NSFC (62466036, 62271239), by High-level and Urgently Needed Overseas Talent Programs of Jiangxi Province (20232BCJ25024), and by Jiangxi Double Thousand Plan (JXSQ2023201022).

\bibliographystyle{IEEEtran}
\bibliography{refs}

\appendix
\onecolumn

\section*{\textbf{\large{Appendix}}}

\subsection{Dataset and Implementation Details}
We create the largest PD facial expression (PDFE) dataset comprising neutral and 6 basic emotional images from 145 patients along with their PD severity labels (i.e., early-stage / mid\&late-stage) by collaborating with the affiliated hospital.
Informed consents of data collection and usage have been obtained from all participants.
Besides, images from three public facial expression datasets (i.e., RaFD [1], Oulu-CASIA [2], and TFED [3]) are utilized as non-PD samples. 
Training uses the Adam optimizer (lr=0.001, 100 epochs), with $\gamma$ in (2) set at 2, following [4].

For experiments, we evaluate the performance of PD severity diagnosis via 5-fold cross-validation and adopt the average area under curve (AUC), G-Mean, F1-Score, and accuracy as criterion.
 
\subsection*{\textnormal{\textbf{References}}}
 \hangafter 1
 \hangindent 2.5em
[1] O. Langner, R. Dotsch, G. Bijlstra, D. H. Wigboldus, S. T. Hawk, and A. Van Knippenberg, “Presentation and validation of the radboud faces database,” \textit{Cogn. Emot.}, vol. 24, pp. 1377–1388, 2010.

 \hangafter 1
 \hangindent 2.5em
[2] G. Zhao, X. Huang, M. Taini, S. Z. Li, and M. Pietikäinen, “Facial expression recognition from near-infrared videos,” \textit{Image Vis. Comput.}, vol. 29, pp. 607–619, 2011.

 \hangafter 1
 \hangindent 2.5em
[3] T. Yang, Z. Yang, G. Xu, D. Gao, Z. Zhang, H. Wang, S. Liu, L. Han, Z. Zhu, Y. Tian et al., “Tsinghua facial expression database-a database of facial expressions in chinese young and older women and men: Development and validation,” \textit{PLoS One}, vol. 15, no. 4, p. e0231304, 2020.

 \hangafter 1
 \hangindent 2.5em
[4] T.-Y. Lin, P. Goyal, R. Girshick, K. He, and P. Dollár, “Focal loss for dense object detection,” in \textit{ICCV}, 2017, pp. 2980–2988.

\end{document}